\documentclass[letterpaper]{article} 
\usepackage{aaai2026}  
\usepackage{times}  
\usepackage{helvet}  
\usepackage{courier}  
\usepackage[hyphens]{url}  
\usepackage{graphicx} 
\usepackage{multirow}
\usepackage{amsmath}
\usepackage{booktabs}
\usepackage{amsfonts}
\usepackage{natbib}  
\usepackage{caption} 
\usepackage{algorithm}
\usepackage{algorithmic}

\usepackage{newfloat}
\usepackage{listings}
\usepackage{xcolor}

\DeclareCaptionStyle{ruled}{labelfont=normalfont,labelsep=colon,strut=off} 
\floatstyle{ruled}
\newfloat{listing}{tb}{lst}{}
\floatname{listing}{Listing}
\title{MotionFlux: Efficient Text-Guided Motion Generation through\\Rectified Flow Matching and Preference Alignment}
\author{
  Zhiting Gao\textsuperscript{\rm 1}\quad
  Dan Song\textsuperscript{\rm 1}\thanks{Corresponding author.}\quad
  Diqiong Jiang\textsuperscript{\rm 2}\quad
  Chao Xue\textsuperscript{\rm 3}\quad
  An-An Liu\textsuperscript{\rm 1}\footnotemark[1]
}
\affiliations{
  \textsuperscript{\rm 1}Tianjin University, China\\
  \textsuperscript{\rm 2}China University of Petroleum, China\\
  \textsuperscript{\rm 3}Tiandy Technologies, China\\
  \texttt{dan.song@tju.edu.cn, anan0422@gmail.com} \ (Corresponding Authors)
}

\usepackage{bibentry}

\begin{document}
\twocolumn[{
\renewcommand\twocolumn[1][]{#1}
\maketitle
\begin{center}
    \captionsetup{type=figure}
    \includegraphics[width=0.9\textwidth]{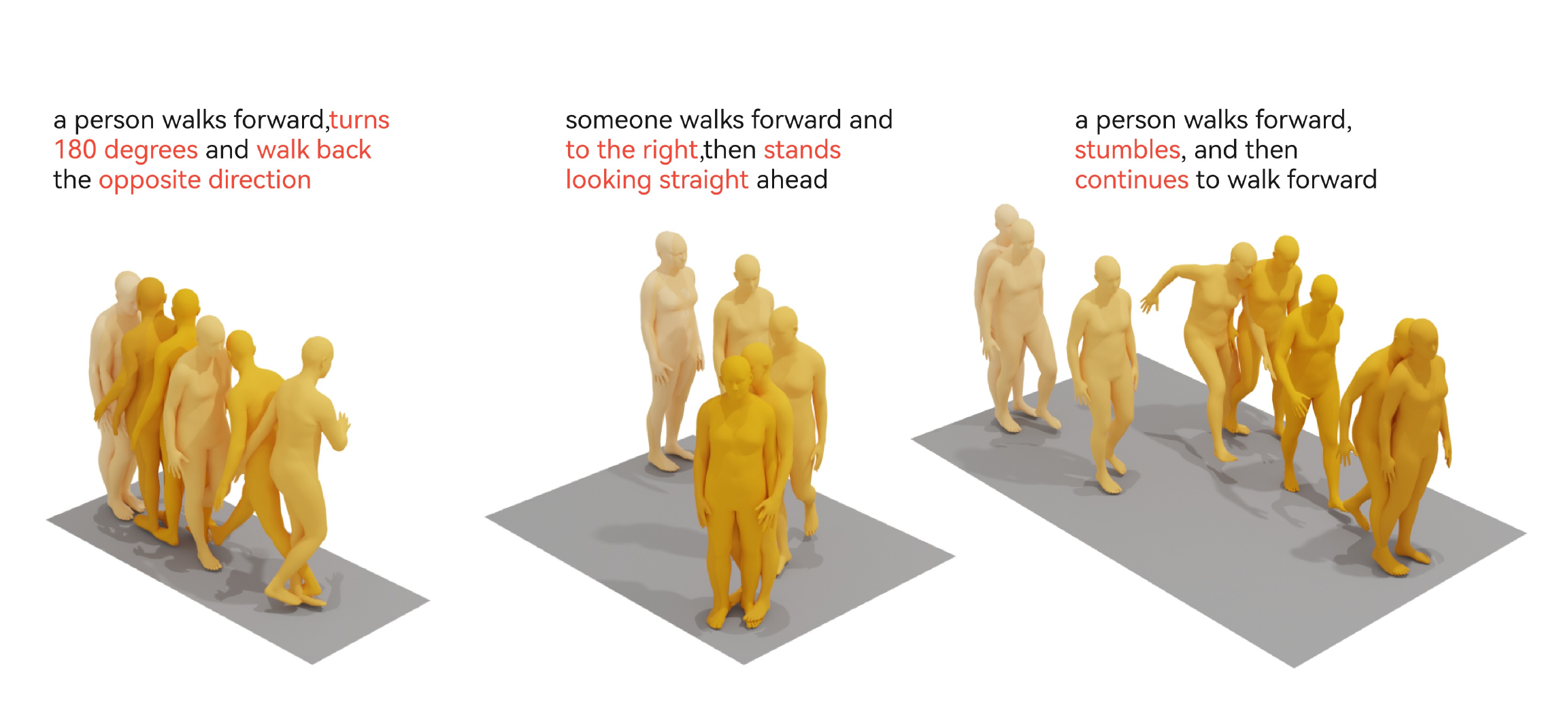}
    \captionof{figure}{We propose MotionFlux, a rectified flow matching-based motion generation framework that employs preference optimization for semantic alignment. In our visualization, darker colors denote later times, and red text highlights key events.}
\end{center}
}]
\begin{abstract}
\indent Motion generation is essential for animating virtual characters and embodied agents. While recent text-driven methods have made significant strides, they often struggle with achieving precise alignment between linguistic descriptions and motion semantics, as well as with the inefficiencies of slow, multi-step inference. To address these issues, we introduce TMR++ Aligned Preference Optimization (TAPO), an innovative framework that aligns subtle motion variations with textual modifiers and incorporates iterative adjustments to reinforce semantic grounding. To further enable real-time synthesis, we propose MotionFLUX, a high-speed generation framework based on deterministic rectified flow matching. Unlike traditional diffusion models, which require hundreds of denoising steps, MotionFLUX constructs optimal transport paths between noise distributions and motion spaces, facilitating real-time synthesis. The linearized probability paths reduce the need for multi-step sampling typical of sequential methods, significantly accelerating inference time without sacrificing motion quality. Experimental results demonstrate that, together, TAPO and MotionFLUX form a unified system that outperforms state-of-the-art approaches in both semantic consistency and motion quality, while also accelerating generation speed. The code and pretrained models will be released.
\end{abstract}


\section{Introduction}
\label{sec:intro}
Natural language-driven motion generation presents significant challenges due to the inherent modality gap between linguistic descriptions and kinematic motions.To address these challenges, existing approaches typically learn precise mappings from language space to motion space through various latent representation learning frameworks. Early attempts employ Auto-Encoders \cite{ahuja2019language2pose,ghosh2021synthesis,tevet2022motionclip} and Variational Auto-Encoders (VAEs) \cite{petrovich2021action,petrovich2022temos} to establish joint embeddings between text and motion. Building on these foundations, notable works include MotionClip \cite{tevet2022motionclip} aligning motion space with CLIP \cite{radford2021learning} embeddings, while ACTOR \cite{petrovich2021action} and TEMOES \cite{plappert2016kit} propose transformer-based VAEs for action-to-motion and text-to-motion tasks respectively. While these methods demonstrate competence with simple descriptions, their performance degrades significantly when handling complex, lengthy textual prompts.

\begin{figure*}[htbp]
  \centering
  \includegraphics[width=\textwidth]{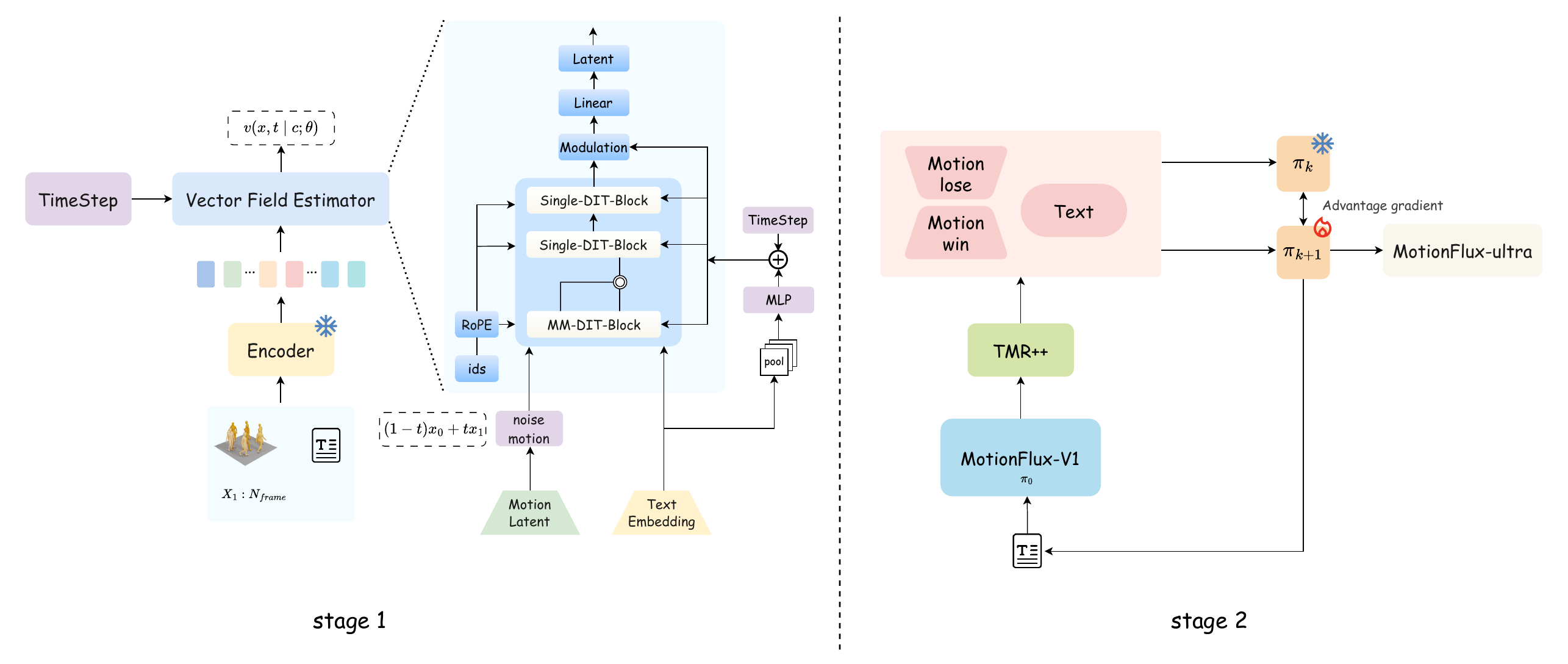}
  \caption{\textbf{Overview of MotionFlux}. In the first stage, we begin by utilizing a pre-trained VAE (with frozen parameters) to compress the raw motion sequence \( X_{1:N_{\text{frame}}} \) into the latent space. The compressed representation, along with the text embedding and timestep, is then fed into the vector estimator to obtain the vector field prediction \( v \). In the second stage, we freeze the model parameters trained in the first stage and use them as a reference model. We then select random texts from the initial training set to generate an online dataset. The optimization objective in the second stage is to iteratively generate high-quality outputs that align with the expected targets by comparing the predictions of the main model and the reference model.}
  \label{fig1}
\end{figure*}

Recent efforts by Guo et al. \cite{guo2022generating} and TM2T \cite{guo2022tm2t} attempt to address intricate text descriptions through multi-stage generation pipelines. However, these approaches suffer from non-intuitive architectures requiring three-phase processing pipelines, often failing to produce high-quality motions that strictly adhere to textual specifications. The emergence of diffusion models \cite{ho2020denoising}, particularly their successful adaptation to motion generation in MDM \cite{tevet2022human} and MotionDiffuse \cite{zhang2024motiondiffuse}, has established diffusion-based methods as state-of-the-art for text-to-motion tasks due to their superior distribution modeling capabilities. Nevertheless, these methods require extensive sampling steps (e.g., 24s for MDM \cite{tevet2022human} and 14s for MotionDiffuse \cite{zhang2024motiondiffuse} per sequence) even with acceleration techniques \cite{song2020denoising}, severely limiting their applicability in real-time scenarios. 
To overcome the slow inference inherent in diffusion-based models, we introduce MotionFLUX, a novel high-speed text-to-motion generation framework based on rectified flow matching. Unlike conventional diffusion models that require hundreds of iterative denoising steps, MotionFLUX learns a deterministic mapping of motion trajectories in latent space, enabling generation in a single or just a few steps. A key advantage of rectified flow is its ability to learn a global, time-consistent velocity field, effectively avoiding the step-wise error accumulation common in diffusion processes. As a result, MotionFLUX achieves substantial improvements in inference speed—crucial for real-time and interactive scenarios—while maintaining or even surpassing the motion quality of state-of-the-art diffusion models.

Moreover, current text-to-motion systems face critical alignment challenges. A central challenge in achieving alignment for text-to-motion models is the creation of preference pairs. Unlike LLM alignment, where established reward models\cite{lambert2024t} and human feedback data or verifiable gold-standard answers are available, alternatively, in the text-to-audio field\cite{hung2024tangoflux}, CLAP\cite{wu2023large} can be introduced as a reward model, the text-to-motion field currently lacks such resources. For instance, leading-edge LLMs, like GPT-4 \cite{achiam2023gpt}, are often directly used to evaluate candidate outputs \cite{zheng2023judging}. Although video-language models\cite{zhang2023video} can process action video inputs and generate text outputs, they typically produce noisy feedback, which is not suitable for generating action preference pairs. If human annotators were to assign binary scores to each video sample based on its alignment with a given prompt, this would quickly become economically unfeasible in large-scale applications.

To overcome this, we introduce TAPO (TMR++ Aligned Preference Optimization), a lightweight powerful alignment framework as depicted in Figure \ref{fig1}. TAPO leverages an internal scoring mechanism (TMR++) to automatically compare candidate motion-text pairs and construct preference data—identifying which generated motion better matches the given prompt. This enables an online, self-supervised optimization loop that continuously refines the model without requiring manual annotation or external reward models. By integrating TAPO into our pipeline, we achieve a much stronger alignment between generated motions and complex language instructions.

\noindent Our main contributions are threefold:

\begin{itemize}
\item \textbf{MotionFLUX - The first high-speed text-to-motion model leveraging rectified flow matching.}
It innovatively introduces rectified flow matching to text-driven motion generation, overcoming slow sampling limitations of diffusion models. It achieves high-quality generation with very few steps, greatly improving inference speed and enabling real-time applications.
\item \textbf{TAPO - An automatic online preference learning framework for motion-text alignment.} We proposes a novel self-supervised paradigm that uses TMR++ as a surrogate reward model to automatically generate preference pairs. This addresses the challenge of costly human annotations and noisy feedback, enabling continuous self-improvement of alignment quality between complex text semantics and motion.
\item Extensive experiments showing MotionFLUX's superior balance between motion quality, text alignment, and real-time efficiency across multiple evaluation protocols.
\end{itemize}

\section{Related Work}
\label{sec:formatting}
\noindent\textbf{Human Motion Synthesis.} Human motion synthesis has been widely explored in computer vision and graphics, with early work focusing on deterministic motion prediction using RNNs \cite{butepage2017deep}, GANs \cite{barsoum2018hp}, GCNs \cite{mao2019learning}, and MLPs \cite{bouazizi2022motionmixer}. To enhance diversity, VAEs \cite{aliakbarian2020stochastic} were introduced. Tasks like motion in-betweening \cite{duan2021single} and trajectory-based generation \cite{pavllo2018quaternet} expanded the scope of motion synthesis. Methods for animator control, including convolutional autoencoders \cite{holden2016deep} and phase-functioned networks \cite{holden2017phase}, were also developed within the graphics community.

Recent advancements in large-scale datasets and multi-task frameworks have driven further progress in text-conditioned motion synthesis. The HumanML3D dataset \cite{guo2022generating} and TM2T \cite{guo2022tm2t} enabled more effective training and performance improvements. Diffusion models, such as MDM \cite{tevet2022human} and MotionDiffuse \cite{yan2018mt}, have led to breakthroughs in both motion quality and diversity. Despite these advances, aligning generated motions with complex textual semantics, particularly for out-of-distribution descriptions, remains challenging. This work addresses this issue by using Rectified Flow Matching to achieve fine-grained semantic alignment, high-quality generation, and real-time synthesis for text-to-motion tasks.

\noindent\textbf{Preference Optimization.} Preference optimization is widely recognized as a key method for aligning large language models (LLMs), either by training a dedicated reward model to encode human preferences \cite{ouyang2022training} or by repurposing the LLM itself as the reward model \cite{rafailov2023direct}. More recent work refines this process through iterative alignment, leveraging human annotators to form preference pairs or relying on pre-trained reward models \cite{kim2024sdpo,gulcehre2023reinforced}. Incorporating verifiable answers can further facilitate the construction of preference pairs. Beyond LLMs, Diffusion-DPO\cite{wallace2024diffusion} demonstrates that diffusion and flow-based models can be aligned similarly. However, aligning Text-to-motion poses unique challenges, as there is no definitive “gold” motion for each text prompt and motion perception remains subjective.Therefore, preference optimization research for text-to-motion generation remains relatively limited.

\section{Method}
MotionFLUX is designed for motion synthesis by integrating FluxTransformer-based architectures with a two-stage preference-driven training pipeline. The model leverages latent motion embeddings extracted via a Variational Autoencoder (VAE) \cite{kingma2013auto} and combines the Diffusion Transformer (DiT) framework \cite{peebles2023scalable} with its multimodal extension MMDiT \cite{esser2024scaling}. Figure\ref{fig1} provides an overview of the MotionFLUX architecture and training workflow. Our approach follows Two-Stage optimization pipeline:
\begin{itemize}
    \item \textbf{Representation Learning Stage}: Our approach concentrates on acquiring essential knowledge from large-scale motion datasets through a variational autoencoder and transformer-based encoders.
    \item \textbf{Preference Alignment Stage}: We refine the model with the TMR++ Aligned Preference Optimization (TAPO) framework, which generates synthetic motion samples and applies comparative preference rankings to enhance semantic alignment and motion quality.
\end{itemize}

\subsection{MODEL CONDITIONING}
\noindent\textbf{Motion Encoder.} In the MotionFLUX model, Motion Encoding captures temporal patterns in motion sequences using a Variational Autoencoder (VAE) with Transformer-based architectures. Given a motion sequence \(X \in \mathbb{R}^{n \times j}\) where \(n\) denotes the number of frames for the action duration and \(j\) represents the spatial positions of the human motion tree structure. The input motion are embedded into a latent space through a linear transformation and appends a global motion token to represent overarching motion patterns across frames. A Transformer encoder then produces a latent distribution parameterized by 
mean $\mu$ and log variance 
$\sigma$, from which we sample a latent motion representation \(Z \in \mathbb{R}^{L \times C}\) using the reparameterization trick.
where \(L\) and \(C\) denote the latent sequence length and the number of channels, respectively. This encoding approach effectively models complex temporal and spatial motion dependencies, providing robust representations for tasks such as motion prediction and generation.

\noindent\textbf{Textual  Conditioning.} Given the textual description of a motion, we obtain the text encoding \textbf{c\textsubscript{text}} from a pretrained text-encoder. Considering the strong performance of FLAN-T5 \cite{chung2024scaling} as conditioning in text-to-motion generation \cite{dai2024motionlcm}, we select FLAN-T5\cite{chung2024scaling} as our text encoder.

\subsection{MODEL ARCHITECTURE}
MotionFLUX adopts a hybrid Transformer backbone inspired by recent FLUX models for image generation \footnote{\url{https://blackforestlabs.ai/}}. MotionFLUX integrates one MMDiT block for robust multimodal fusion and two DiT blocks for efficient temporal reasoning. Replacing part of MMDiT with DiT improves parameter efficiency while maintaining strong performance \footnote{\url{https://blog.fal.ai/auraflow/}}.  Each block uses 6 attention heads (head dimension 128, hidden width 768), resulting in a 43M-parameter model that balances scalability and expressiveness.

\subsection{Flow Matching}
Flow Matching (FM) 
\cite{lipman2022flow, albergo2022building} provides a robust alternative to diffusion-based generative models \cite{ho2020denoising, song2020denoising}, which are highly sensitive to noise scheduler. FM learns a time-dependent vector field that transports samples from a simple prior (e.g., Gaussian) to a complex target distribution.

\noindent\textbf{Rectified Flows}. In our framework, we employ rectified flows \cite{liu2022flow}, which define the shortest linear transport path from noise to the target distribution.
Let \(x_1\) denote the latent representation of a motion sample, and let \(x_0 \sim \mathcal{N}(0, \mathbf{I})\) be a noise sample. For a given time step \(t \in [0, 1]\), we construct a training sample \(x_t\) so that the model learns to predict a velocity \(v_t = \frac{\mathrm{d}x_t}{\mathrm{d}t}\) that drives \(x_t\) towards \(x_1\). Although there are several strategies to construct the transport path \(x_t\), we adopt rectified flows\cite{liu2022flow}, in which the forward process follows straight-line paths connecting the target and noise distributions, as defined in Eq.\ref{eq:1}. Empirical results have shown that rectified flows are sample-efficient and experience less degradation than alternative formulations when fewer sampling steps are used \cite{esser2024scaling}. We denote the parameters of the model \(u\) by \(\theta\). The model is trained by directly regressing the predicted velocity $v(x, t \mid c; \theta)$ against the ground truth velocity \(v_t\) using the loss function in Eq.\ref{eq:2}.

\begin{equation}
x_t = (1 - t)x_1 + t{x}_0, \quad v_t = \frac{dx_t}{dt} = {x}_0 - x_1,
\label{eq:1}
\end{equation}
\begin{equation}
\mathcal{L}_{FM} = \mathbb{E}_{x_1, x_0, t} \left\| v(x, t \mid c; \theta) - v_t \right\|^2,
\label{eq:2}
\end{equation}

\begin{figure}[t]
  \centering
  \includegraphics[width=0.6\linewidth]{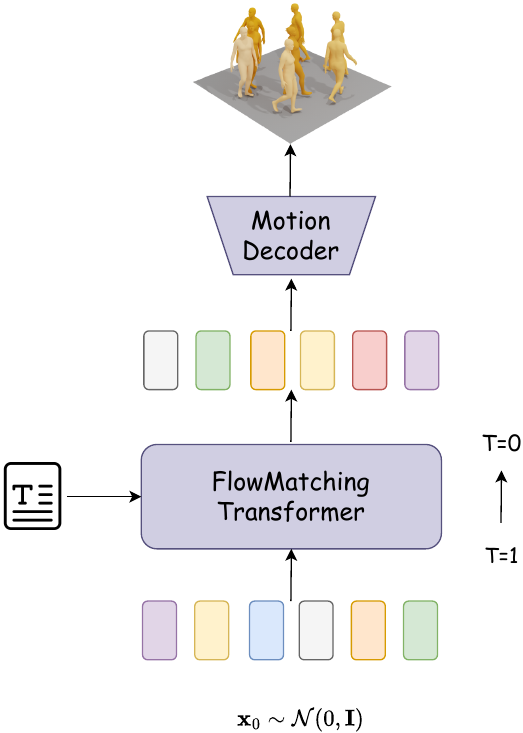}
  \caption{Overview of the sampling pipeline employed in our rectified-flow–based text-to-motion framework}
  \label{fig2}
\end{figure}

\noindent\textbf{Inference}.During inference, a noise sample \({x}_0 \sim \mathcal{N}(0, \mathbf{I})\) is drawn from the prior, and an ordinary differential equation solver is used to compute \(x_1\) based on the predicted velocity of the model \(v_t\) at each time step \(t\). In our implementation, we utilize an Euler solver for this purpose:
\begin{equation}
x_{t+\epsilon} = x + \epsilon\, v(x,t \mid c;\theta),
\end{equation}
where $\epsilon$ is the step size, the latent representation obtained from sampling is then passed to the motion decoder to reconstruct the motion. Figure \ref{fig2} illustrates the overall sampling process of our model.

\subsection{TMR++ ALIGNED PREFERENCE OPTIMIZATION}

\noindent\textbf{TMR++ Aligned Preference Optimization (TAPO)} aligns the MotionFLUX model with textual intent through iterative, preference-based training. It leverages \textbf{TMR++ (Text-to-Motion Retrieval)} \cite{bensabath2024cross} as a proxy reward model to evaluate generated motions according to their semantic consistency with the input text and to construct preference pairs for direct optimization.

TMR++ is a contrastive learning–based cross-modal retrieval framework that jointly trains text-to-motion generation and retrieval tasks. It optimizes a shared text–motion embedding space using the InfoNCE loss and applies a text-similarity–driven negative sampling strategy to enhance representation quality. 

In TAPO, the reward score is computed as the cosine similarity between the textual embedding and the generated motion embedding, providing a quantitative measure of semantic alignment. 
As shown in Section~4.3, using TMR++ to perform best-of-$N$ candidate selection improves both objective metrics and motion quality.

The TAPO alignment process begins with a pretrained MotionFLUX checkpoint $\pi_0$ as the base policy. 
It then iteratively updates the policy $\pi_k = u(\cdot; \theta_k)$ to $\pi_{k+1}$ through three steps: (1) batched online data generation, (2) reward estimation and preference dataset creation, and (3) fine-tuning $\pi_k$ into $\pi_{k+1}$ via direct preference optimization.

Our approach is inspired by LLM alignment techniques \cite{zelikman2022star,kim2024sdpo} but differs in two key aspects:
\begin{enumerate}
    \item It focuses on \emph{rectified-flow motion generation} rather than autoregressive language modeling.
    \item The motion domain lacks \emph{off-the-shelf reward models} that are commonly used in LLM alignment to produce reliable preference datasets.
\end{enumerate}

By leveraging TMR++ as a proxy reward model and performing online preference data generation at each iteration, TAPO avoids preference saturation, maintains training stability, and progressively enhances semantic alignment and motion quality.

\subsection{DYNAMIC ONLINE BATCH DATA GENERATION}
During the construction of the \(k\)th iteration preference dataset, we first randomly sample a subset \(\mathcal{M}_k\) from the large-scale prompt pool \(\mathcal{B}\) to serve as the text input for the current batch. For each text prompt \(y_i \in \mathcal{M}_k\), the current policy \(\pi_k\) generates \(N\) candidate motion samples, which are then ranked according to semantic similarity using the TMR++ \cite{bensabath2024cross} cross-modal evaluation framework. Specifically, by computing the semantic consistency score between the generated motion and the original text prompt \(y_i\), the sample with the highest score is designated as the superior sample \(x_i^w\) and the one with the lowest score as the inferior sample \(x_i^l\). The resulting preference dataset \(\mathcal{D}_k\) is composed of triples \((x_i^w, x_i^l, y_i)\), with each triple corresponding to a specific text prompt in \(\mathcal{M}_k\).This process ensures data diversity through a dynamic sampling mechanism and achieves objective selection with the aid of quantitative evaluation tools. It is important to emphasize that the cross-modal alignment capability of the TMR++ \cite{bensabath2024cross} framework plays a crucial role in this process, as its fine-grained similarity measurement effectively captures the semantic correlation between text and motion. This contrastive selection-based data construction strategy not only preserves the core idea of the original algorithm but also enhances the interpretability of the method through hierarchical sampling and quantitative evaluation.

\subsection{PREFERENCE OPTIMIZATION}
Direct preference optimization (DPO) \cite{rafailov2023direct} has proven effective in encoding human preferences into large language models (LLMs) \cite{ouyang2022training}. Building on this success, DPO has been adapted into the DPO-Diffusion framework \cite{wallace2024diffusion} for aligning diffusion models. The DPO-Diffusion loss is defined as:

\begin{equation}
	\begin{aligned}
		L_{\mathrm{DPO-Diff}}
		&=-\mathbb{E}_{n,\epsilon^w,\epsilon^l}\log\sigma\left(-\beta(\|\epsilon_n^w-\epsilon_\theta(x_n^w)\|_2^2 \right.\\
		&\quad -\|\epsilon_n^w-\epsilon_{\mathrm{ref}}(x_n^w)\|_2^2-(\|\epsilon_n^l-\epsilon_\theta(x_n^l)\|_2^2\\
		&\quad \left.-\|\epsilon_n^l-\epsilon_{\mathrm{ref}}(x_n^l)\|_2^2))\right),
        \label{eq:4}
	\end{aligned}
\end{equation}
where the loss compares the noise-prediction errors of winning and losing  samples during the diffusion process. By applying a sigmoid function with a scaling factor $\beta$, it reinforces the model’s capacity to capture human preferences, thereby ensuring effective alignment of the diffusion model.

As demonstrated by \cite{wallace2024diffusion}, through some algebraic manipulation the expression in Eq.\ref{eq:4} can be simplified into the more tractable form presented in Eq.\ref{eq:5}. In this setting, \(T\) represents the diffusion timestep with \(t\) uniformly sampled from \(U(0,T)\); \(x_n^w\) and \(x_n^l\) denote the winning and losing motions, respectively; and the noise term \(\epsilon\) follows the distribution \(\mathcal{N}(0,\mathbf{I})\).
Following \cite{esser2024scaling}, the DPO-Diffusion loss can be extended to rectified flows. By exploiting the equivalence \cite{lipman2022flow} between \(\epsilon_\theta\) and \(u(\cdot;\theta)\), the noise matching terms can be replaced with flow matching terms, resulting in

\begin{footnotesize}
 \begin{align}
  L_{\mathrm{DPO-FM}}&=-\mathbb{E}_{t\sim\mathcal{U}(0,1),x^w,x^l}\log\sigma\{-\beta(\underbrace{\|u(x_t^w,t;\theta)-v_t^w\|_2^2}_{\text{Winning loss}} \notag \\
  & \quad -\underbrace{\|u(x_t^l,t;\theta)-v_t^l\|_2^2}_{\text{Losing loss}})-(\underbrace{\|u(x_t^w,t;\theta_{\mathrm{ref}})-v_t^w\|_2^2}_{\text{Winning reference loss}} \notag \\
  & \quad -\underbrace{\|u(x_t^l,t;\theta_{\mathrm{ref}})-v_t^l\|_2^2}_{\text{Losing reference loss}})\},
  \label{eq:5} 
 \end{align}
\end{footnotesize}

\noindent
where \(t\) denotes the flow matching timestep, and \(x_t^l\) along with \(x_t^w\) correspond to the losing and winning motions, respectively.

\medskip

In the context of LLMs, the DPO loss models the relative likelihood between winning and losing responses. Minimizing this loss effectively increases the margin between them—even if both log-likelihoods are reduced \cite{pal2024smaug}. Since DPO emphasizes the relative likelihood rather than the absolute values, convergence requires that both likelihoods decrease, which might seem counterintuitive \cite{rafailov2024r}. Notably, this reduction in likelihood does not inherently harm performance; rather, it is a necessary condition for improvement \cite{rafailov2024scaling}. When applied to rectified flows, however, the situation becomes more complex due to the challenges in estimating the likelihood of samples generated under classifier-free guidance (CFG). A closer inspection of \(L_{\text{DPO-FM}}\) Eq. \ref{eq:5} reveals that minimizing it can be achieved by simply enlarging the gap between the winning and losing losses, even if both losses increase. 
\medskip

To address this issue, we propose incorporating the winning loss directly into the optimization objective, thereby preventing its inadvertent increase. Our modified loss is defined as
\begin{equation}
L_{\text{TAPO}} = L_{\text{DPO-FM}} + \alpha L_{\text{FM}},
\label{eq:6}
\end{equation}
where $L_{\text{FM}}$ is the flow matching loss computed exclusively on the winning motion, as given in Eq.~\ref{eq:6}, and $\alpha$ is a learnable parameter balancing its influence. 

Although the DPO loss effectively improves preference ranking between chosen and rejected motions, relying solely on it can cause overoptimization. This may compromise the semantic and structural fidelity of the winning motion, leading the model’s outputs to drift away from the intended distribution. By adding the $L_{\text{FM}}$ term, we anchor the model to the high-quality attributes of the winning examples, regularizing the training process. This additional loss stabilizes optimization and helps preserve the essential characteristics of the winning motions, resulting in a more balanced and robust performance.

\begin{table*}[t]
    \centering
    \caption{Comparison of text-conditional motion synthesis on the HumanML3D \cite{guo2022generating} dataset. We compute the suggested metrics following \cite{guo2022generating}. The evaluation is repeated 20 times for each metric and the average is reported with a 95\% confidence interval. ``$\rightarrow$'' indicates that the closer to the real data, the better. Bold and underlined entries indicate the best and the second best results, respectively. The MotionFlux-ultra(5ms) surpasses all state-of-the-art models.}
    \label{tab:humanml3d_comparison}
    \resizebox{\textwidth}{!}{%
    \begin{tabular}{l c c c c c c c c}
    \toprule
    \multirow{2}{*}{\textbf{Methods}} 
    & \multirow{2}{*}{\textbf{AITS}($\downarrow$)} 
    & \multicolumn{3}{c}{\textbf{R-Precision}($\uparrow$)} 
    & \multirow{2}{*}{\textbf{FID} ($\downarrow$)} 
    & \multirow{2}{*}{\textbf{MM Dist} ($\downarrow$)} 
    & \multirow{2}{*}{\textbf{Diversity} ($\rightarrow$)} 
    & \multirow{2}{*}{\textbf{MultiModality} ($\uparrow$)}\\
    \cmidrule(lr){3-5}
    & & Top 1 & Top 2 & Top 3 &  &  &  &  \\
    \midrule
    Real & - & 0.511$^{\pm .003}$ & 0.703$^{\pm .003}$ & 0.797$^{\pm .002}$ & 0.002$^{\pm .000}$ & 2.794$^{\pm .008}$ & 9.503$^{\pm .065}$ & -\\
    \midrule
    Seq2Seq \cite{lin2018generating} & -         & 0.180$^{\pm .002}$ & 0.300$^{\pm .002}$ & 0.396$^{\pm .002}$ & 11.75$^{\pm .035}$ & 5.529$^{\pm .007}$ & 6.223$^{\pm .061}$ & - \\
    JL2P \cite{ahuja2019language2pose}    & -         & 0.246$^{\pm .002}$ & 0.387$^{\pm .002}$ & 0.486$^{\pm .002}$ & 11.02$^{\pm .046}$ & 5.296$^{\pm .008}$ & 7.676$^{\pm .058}$ & - \\
    T2G \cite{bhattacharya2021text2gestures}     & -         & 0.165$^{\pm .001}$ & 0.267$^{\pm .002}$ & 0.345$^{\pm .002}$ & 7.664$^{\pm .030}$  & 6.030$^{\pm .008}$ & 6.409$^{\pm .071}$ & - \\
    Hier \cite{ghosh2021synthesis}   & -         & 0.301$^{\pm .002}$ & 0.425$^{\pm .002}$ & 0.552$^{\pm .004}$ & 6.532$^{\pm .024}$  & 5.012$^{\pm .018}$ & 8.332$^{\pm .042}$ & - \\
    TEMOS \cite{petrovich2022temos}  & 0.017     & 0.424$^{\pm .002}$ & 0.612$^{\pm .002}$ & 0.722$^{\pm .002}$ & 3.734$^{\pm .028}$  & 3.703$^{\pm .008}$ & 8.973$^{\pm .071}$ & 0.368$^{\pm .018}$ \\
    T2M \cite{guo2022generating}    & 0.038     & 0.457$^{\pm .002}$ & 0.639$^{\pm .003}$ & 0.740$^{\pm .003}$ & 1.067$^{\pm .002}$  & 3.340$^{\pm .008}$ & 9.188$^{\pm .002}$ & 2.090$^{\pm .083}$ \\
    MDM \cite{tevet2022human}    & 24.74     & 0.320$^{\pm .005}$ & 0.498$^{\pm .004}$ & 0.611$^{\pm .007}$ & 0.544$^{\pm .044}$  & 5.566$^{\pm .027}$ & 9.559$^{\pm .086}$ & \textbf{2.799}$^{\pm .072}$ \\
    MotionDiffuse \cite{zhang2024motiondiffuse} & 14.74 & 0.491$^{\pm .001}$ & 0.681$^{\pm .001}$ & 0.782$^{\pm .001}$ & 0.630$^{\pm .001}$  & 3.113$^{\pm .001}$ & 9.410$^{\pm .049}$ & 1.553$^{\pm .042}$ \\
    MLD \cite{chen2023executing}     & 0.217     & 0.481$^{\pm .003}$ & 0.673$^{\pm .003}$ & 0.772$^{\pm .002}$ & 0.473$^{\pm .013}$  & 3.196$^{\pm .010}$ & 9.724$^{\pm .082}$ & \underline{2.413}$^{\pm .079}$ \\
    MotionLCM\cite{dai2024motionlcm}  & 0.030 & 0.502$^{\pm .003}$ & 0.701$^{\pm .002}$ & 0.803$^{\pm .002}$ & 0.467$^{\pm .012}$ & 3.022$^{\pm .009}$ & 9.631$^{\pm .066}$ & 2.172$^{\pm .082}$ \\
    \midrule
    MotionFlux-V1 
    & $\textbf{0.005}$
    & $\underline{0.530}^{\pm 0.002}$ & $\underline{0.729}^{\pm 0.001}$ & $\underline{0.825}^{\pm 0.001}$ 
    & $\underline{0.086}^{\pm 0.003}$ & $\underline{2.87}^{\pm 0.006}$ & $\textbf{9.531}^{\pm 0.09}$ & $1.841^{\pm 0.07}$ \\

    \midrule
    MotionFlux-ultra 
    & $\textbf{0.005}$
    & $\textbf{0.536}^{\pm 0.003}$ & $\textbf{0.732}^{\pm 0.002}$ & $\textbf{0.827}^{\pm 0.001}$ 
    & $\textbf{0.078}^{\pm 0.004}$ & $\textbf{2.84}^{\pm 0.005}$ & $\textbf{9.531}^{\pm 0.09}$ & $1.998^{\pm 0.07}$ \\
    
    \bottomrule
    \end{tabular}}
\end{table*}

\section{Experiment}
In this section, we provide a detailed description of the experimental setup and evaluation process. Section 4.1 covers the datasets, evaluation metrics, model configurations, training procedures, and computational resources used in our experiments. Section 4.2 presents a comparison of our method with competitive approaches. Finally, in Section 4.3, we offer an analysis and discussion of the results.
\subsection{Experimental setup}
\noindent\textbf{Datasets:} We conduct experiments using the widely used HumanML3D dataset \cite{guo2022generating}, which contains 14,616 unique human motion sequences and 44,970 associated textual descriptions. To ensure a fair comparison with prior methods \cite{chen2023executing,guo2022generating,petrovich2022temos,tevet2022human,zhang2024motiondiffuse}, we utilize the redundant motion representation, including root velocity, root height, local joint positions, velocities, rotations in root space, and foot contact binary labels.

\noindent\textbf{Evaluation Metrics:} Building on the evaluation metrics established in previous works\cite{chen2023executing,guo2022generating,xie2023omnicontrol} , we extend these measures to comprehensively assess our model. To quantify time efficiency, we report the Average Inference Time per Sentence (AITS) as in \cite{chen2023executing}. For motion quality, we adopt the Frechet Inception Distance (FID) as the primary metric, which evaluates the similarity between the feature distributions of generated and real motions using the feature extractor from \cite{guo2022generating}. Motion diversity is evaluated by two measures: MultiModality (MModality), which assesses the diversity of generated outputs conditioned on identical textual inputs, and Diversity, which computes the variance of the extracted features \cite{guo2022generating}. Condition matching is gauged by calculating the motion-retrieval precision (R-Precision) for text-motion Top-1/2/3 matching accuracy and the Multimodal Distance (MM Dist), which reflects the mean distance between motion and text embeddings.

\noindent\textbf{Experimental Settings:}
We pre-trained \textbf{MotionFlux} on the HumanML3D dataset \cite{guo2022generating} for 500 epochs using the AdamW optimizer \cite{loshchilov2017decoupled} with $\beta_1=0.9$, $\beta_2=0.999$, and an initial learning rate of $1 \times 10^{-4}$. A linear learning-rate scheduler was applied throughout training, and the model was trained on a single A100 GPU with a batch size of 64.
Following prior findings that sampling timesteps $t$ from the middle of $[0,1]$ improves generation quality \cite{hang2023efficient,kim2024adaptive,karras2022elucidating}, we sampled $t$ from a logit-normal distribution with mean $0$ and variance $1$, following \cite{esser2024scaling}. This pre-trained model is referred to as \texttt{MotionFlux-V1}. During the TAPO alignment phase, we used the same optimizer but with an effective batch size of 32, a peak learning rate of $1 \times 10^{-5}$, and a linear warmup of 100 steps. Each TAPO iteration trained for 8 epochs, and the checkpoint from the last epoch was used for batched online data generation. We performed three TAPO iterations, as performance plateaued beyond that point. The final aligned model is referred to as \texttt{MotionFlux-Ultra}.

\subsection{Comparison with Other Methods}
\textbf{Comparison of Text-Conditional Motion Synthesis.} As shown in Table~\ref{tab:humanml3d_comparison}, MotionFlux achieves the lowest AITS, confirming its superior inference efficiency. 
It also obtains the highest R-Precision and lowest MM Dist, demonstrating strong semantic alignment with textual inputs. 
At the same time, MotionFlux achieves the lowest FID, demonstrating that its outputs remain close to real motion distributions, yet it maintains diversity at a level comparable to real data. This balance shows that MotionFlux enhances alignment and realism without collapsing variability.  
Compared with prior methods, MotionFlux achieves a better trade-off between semantic accuracy, motion fidelity, and variability.

\begin{figure}[t]
  \centering
  \includegraphics[width=1\linewidth]{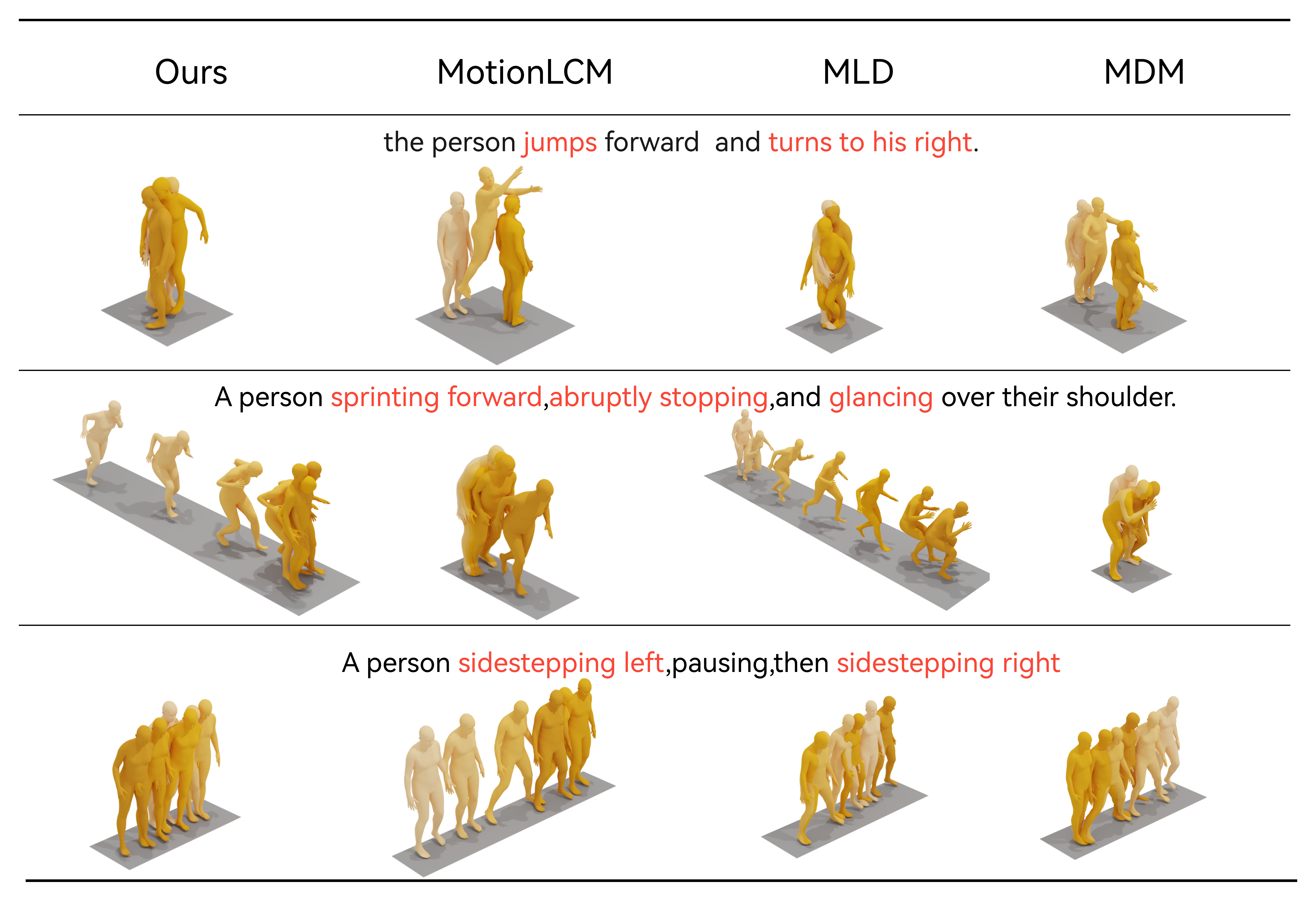}
  \caption{Qualitative comparison of the state-of-the-art methods in the text-to-motiontask,the darker the color, the later the time. We employed ChatGPT-o3 to randomly generate three prompts—none of which had appeared in the dataset—for inference. The visualization results show that MotionFlux exhibits strong semantic alignment on critical events (e.g., “left and right,” “glance”) and demonstrates robust generalization performance.}
  \label{fig_compare}
\end{figure}

\noindent\textbf{Comparison of Fine-Grained Semantic Alignment.} As illustrated in Figure~\ref{fig_compare}, we compare MotionFlux with MotionLCM\cite{dai2024motionlcm}, MLD\cite{chen2023executing}, and MDM\cite{tevet2022human}. MotionLCM frequently produces semantic errors under fine-grained conditions, resulting in reduced motion quality and weaker control. In contrast, MotionFlux consistently achieves higher fidelity and superior alignment. For example, in the second case, MotionLCM and MDM fail to capture the “glance” action, while in the third, both confuse left and right. MotionFlux, however, preserves pose constraints and faithfully follows detailed textual descriptions. In terms of inference speed, MotionFlux is 3 times faster than MotionLCM, 40 times faster than MLD, and 4800 times faster than MDM.

\subsection{Discussion}
\begin{figure}[t]
  \centering
  \includegraphics[width=\linewidth]{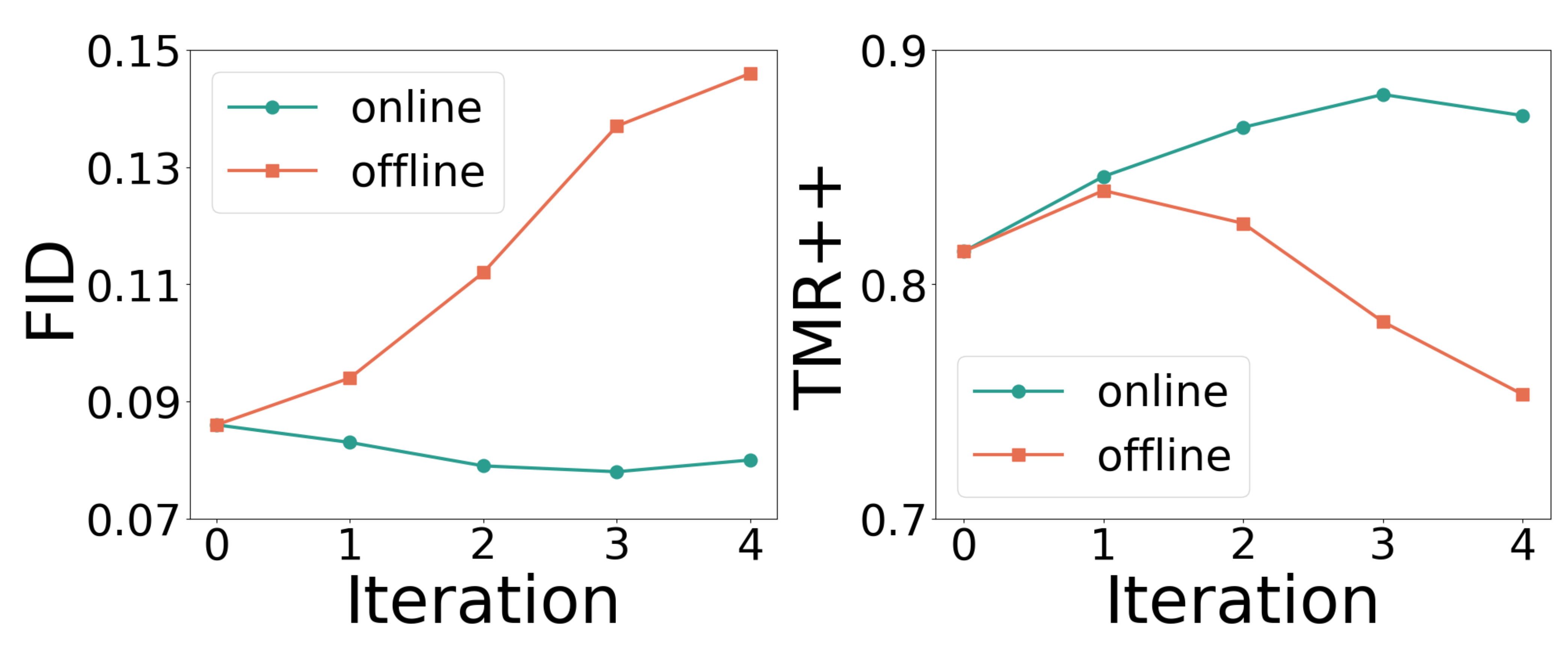}
  \caption{Trajectory of FID and TMR++ scores over training iterations. Offline training peaks by the second iteration with rising FID, while online training continues to improve, showing lower FID and higher TMR++ scores.}
  \label{fig_tmr}
\end{figure}

\noindent\textbf{The Significance of Online Dataset Generation.} Due to the lack of reliable offline data, we resorted to using the output generated by motion-v1 during its first iteration as our offline dataset. Figure \ref{fig_tmr} presents the results of four iterations comparing scenarios with and without new data generation. Our experiments indicate that repeatedly training on a fixed offline dataset leads to rapid performance saturation and subsequent deterioration. Specifically, for the offline TAPO, the TMR++ score begins to decline by the second iteration while the fid metric rises sharply, and by the fourth iteration the model’s performance has substantially degraded---highlighting the drawbacks of relying on offline data. In contrast, the online TAPO, which generates new data at each iteration, consistently outperforms its offline counterpart on both the TMR++ score and fid metrics.
A plausible explanation for this phenomenon is reward over-optimization \cite{rafailov2024scaling}. In the context of DPO training for large language models, \cite{kim2024sdpo} demonstrated that the reference model functions as a sampling lower bound; iteratively updating this reference with the same dataset can cause the current model to minimize the loss in unexpected ways. Furthermore, the true objective of our model is to train a model that is jointly endorsed by both the reward proxy and the reference model (i.e., the model being optimized). The objective function is formulated as follows:
\[
\pi^*(y \mid x) 
= \frac{1}{Z(x)} \,
  \pi_{\text{ref}}(y \mid x) \,
  \exp\!\biggl(\frac{1}{\beta}\,r'(x,y)\biggr),
\]
where \(z(x)\) denotes a normalization factor, \(\pi_{\text{ref}}(y \mid x)\) is the reference model, \(r'(x,y)\) represents the reward function learned from TMR++ feedback, and \(\beta\) is a hyperparameter that controls the temperature (or, equivalently, the inverse temperature).

\begin{table}[t]
\centering
\begin{tabular}{cccc}
\toprule
\textbf{N} & \textbf{FID} $\downarrow$ & \textbf{Diversity} $\uparrow$ & \textbf{TMR++}$_{\text{score}}$ $\uparrow$ \\
\midrule
1 & 0.102 & 9.635 & 0.803 \\
5 & 0.093 & 9.576 & 0.812 \\
10 & 0.086 & 9.531 & 0.819 \\
15 & 0.087 & 9.542 & 0.821 \\
\bottomrule
\end{tabular}
\caption{Comparison of Best-of-N policy. Metrics include Frechet Inception Distance (FID), Diversity and TMR++ score.}
\label{tab:bestofn}
\end{table}

\noindent\textbf{Best-of-N Inferred Policy.} We conducted experiments to verify that the TMR++ model can serve as an effective proxy reward model for evaluating motion outputs. Specifically, we adopt a Best-of-N strategy to assess MotionFlux, with $N \in \{1, 5, 10, 15\}$. 
For each prompt, the generated motion sequences are ranked according to the TMR++ score.  The results in Table~\ref{tab:bestofn} show that increasing $N$ in the Best-of-$N$ strategy consistently improves both the TMR++ score and FID, while diversity remains stable. 
This indicates that TMR++ reliably ranks motion sequences aligned with textual descriptions, and that these quality gains are achieved without sacrificing motion variability, highlighting the method’s ability to balance semantic alignment and motion quality.

\section{Conclusion}
In summary, we have presented MotionFlux—a fast and efficient text-to-motion model that leverages synthetic preference data generated online. By integrating advanced rectified flow matching with preference alignment techniques, MotionFlux not only produces animations that accurately reflect user prompts but also outperforms existing diffusion-based baselines in both quality and speed. Moreover, our extensive experiments demonstrate state-of-the-art performance on benchmark datasets, thereby confirming the practical viability of our approach for real-time applications. Overall, this work lays a solid foundation for future research on efficient and semantically robust motion generation.

\bibliography{aaai2026}

\end{document}